\pdfoutput=1
\documentclass[11pt]{article}

\usepackage{acl}

\usepackage{times}
\usepackage{latexsym}

\usepackage[T1]{fontenc}
\usepackage[utf8]{inputenc}

\usepackage{microtype}

\usepackage{inconsolata}

\usepackage{graphicx}

\usepackage{amsmath}
\usepackage{fancyhdr}

\title{Fine-Tuning LLMs for Report Summarization: Analysis on Supervised and Unsupervised Data}

\author{
 \textbf{Swati Rallapalli},
 \textbf{Shannon Gallagher},
 \textbf{Andrew O. Mellinger},
 \textbf{Jasmine Ratchford},
\\
 \textbf{Anusha Sinha},
 \textbf{Tyler Brooks},
 \textbf{William R. Nichols},
 \textbf{Nick Winski},
 \textbf{Bryan Brown}
\\
\\
Software Engineering Institute, Carnegie Mellon University
\\
}

\begin{document}
\pagestyle{fancy}
\fancyhead{}
\renewcommand{\headrulewidth}{0pt} % no line in header area
\fancyfoot{} % clear all footer fields
\fancyfoot[LE,RO]{\thepage}    
\fancyfoot[RE,LO]{[Distribution Statement A] Approved for public release and unlimited distribution.}

\fancypagestyle{empty}{
\fancyhead{}
\renewcommand{\headrulewidth}{0pt} % no line in header area
\fancyfoot{} % clear all footer fields
\fancyfoot[LE,RO]{\thepage}    
\fancyfoot[RE,LO]{[Distribution Statement A] Approved for public release and unlimited distribution.}
}

\maketitle
\begin{abstract}
We study the efficacy of fine-tuning Large Language Models (LLMs) for the specific task of report (government archives, news, intelligence reports) summarization. While this topic is being very actively researched - our specific application set-up faces two challenges: (i) ground-truth summaries maybe unavailable (e.g., for government archives), and (ii) availability of limited compute power - the sensitive nature of the application requires that computation is performed on-premise and for most of our experiments we use one or two A100 GPU cards. Under this set-up we conduct experiments to answer the following questions. 
First, given that fine-tuning the LLMs can be resource intensive, is it feasible to fine-tune them for improved report summarization capabilities on-premise? Second, what are the metrics we could leverage to assess the quality of these summaries? We conduct experiments on two different fine-tuning approaches in parallel and our findings reveal interesting trends regarding the utility of fine-tuning LLMs. Specifically, we find that in many cases, fine-tuning helps improve summary quality and in other cases it helps by reducing the number of invalid or garbage summaries. 
\end{abstract}

\section{Introduction}
\label{sec:intro}

%\textbf{Problem and Challenges:} 
Large Language Models (LLMs) have shown tremendous promise for a variety of Natural Language Processing (NLP) applications. However, it is resource intensive to train them from scratch (also called pre-training). For example, training the largest Llama model could cost about $2$-$3M$ dollars\footnote{https://www.cnbc.com/2023/03/13/chatgpt-and-generative-ai-are-booming-but-at-a-very-expensive-price.html}\footnote{https://aws.amazon.com/ec2/instance-types/p4/}. Therefore, a lot of applications take the pre-trained foundation models, and fine-tune them or adapt them for slightly different tasks, or on specific datasets for specific purposes~\cite{finetuning_finance, finetune_facts}.  

Fine-tuning techniques can help us leverage the generalization capabilities of the foundation models, while potentially achieving improved accuracy on specific tasks. The advantage of fine-tuning compared to training from scratch also include: (i) reduced training resource requirement (compute, time) and (ii) reduced data requirement. However, in the case of LLMs even fine-tuning can be resource intensive depending on how much data is required to produce measurable impact~\cite{parrot}. 

In this paper, we focus on the task of report summarization, where reports could be government archives, news or intelligence reports.  The use cases include analysis of government data %archives data %such as document analysis for national security, intelligence reporting 
where the model is deployed on-premise to prevent leakage of sensitive information. Our challenges include: (i) some of the reports we analyze in our work (national archives) do not have ground truth summaries, (ii) we work with limited compute resources: one or two A100 GPUs for all of our experiments, which is likely the case with many small teams. 

We discuss
metrics that we can use to evaluate the summaries when ground truth summaries are not available. Next, we evaluate if fine-tuning the LLMs can help improve the summary quality. While there is much work regarding fine-tuning LLMs, comparing with few-shot learning \cite{fewshot} and Retrieval Augmented Generation \cite{rag}, our work focuses on simply studying the feasibility and utility of fine-tuning. Specifically, we focus on smaller LLMs that are feasible to deploy on-premise. 

We study two forms of fine-tuning in parallel on different datasets, one from the National Archives\footnote{https://www.archives.gov/research} that we collect and curate and other previously studied news datasets. We pick these datasets since they are open datasets that we can use to study and publish the work as proxies for sensitive documents. We describe the fine tuning methodology and data further in Section~\ref{sec:tune}.
We then generate summaries from foundation models as well as fine tuned models and evaluate the summaries. We leverage multiple metrics, including the well known ROUGE~\cite{rouge}, BLEU~\cite{bleu_paper}, METEOR~\cite{meteor} and BERTScore~\cite{bertscore_paper}. In addition, we also design heuristics to detect invalid summaries which occurred when using the Llama 7B model. We leverage a metric based on topic modeling and refer to it as Topic Similarity (TS). Each of the metrics has different advantages and disadvantages that we discuss further in Section \ref{sec:metrics}. We use TS in our study as it is (i) interpretable and (ii) more suitable in the case where ground truth summaries may not be available. Finally, we use AlignScore~\cite{alignscore_paper}, for automatic evaluation of factual consistency.

%\noindent \textbf{Contributions:} 
We summarize the contributions as follows:
\begin{itemize}
    \item We collect and curate a dataset of National Archives documents to study summarization approaches using LLMs.
    \item We pick multiple metrics, and design heuristics to understand the quality of summarization algorithms in this set-up when no ground truth summaries and benchmarks may exist for comparison purposes.
    \item We fine-tune two different models, on three different datasets under limited resource availability and compare the summaries generated by the foundation models and the fine tuned models using the above metrics.
\end{itemize}
We will release the above models and datasets to the public for research purposes on request.

\section{Fine-tuning Methodology and Data}
\label{sec:tune}
In this section, we describe our set-up for fine-tuning LLMs for report summarization. 
Pre-trained foundation models maybe fine tuned under different conditions for improved accuracy on specific tasks,
for example, (i) to learn specific context, vocabulary, and grammar used in particular areas of law \cite{llm_law1}, medicine \cite{llm_medicine1}, government, etc. (via Knowledge Fine-tuning (KFT)), or (ii) to learn to perform certain tasks like summarization, or sentiment classification better (via Format Fine-tuning (FFT)). We study both of these approaches in this paper.
For efficiency of fine-tuning we use Parameter-Efficient Fine-Tuning (PEFT)~\cite{peft} and Deepspeed~\cite{deepspeed} based on the computational requirement of the experiment. With the below optimizations we are able to fine-tune modest sized foundation LLMs (e.g. 7B prameters) on our local compute. 

\noindent \textbf{PEFT:} PEFT methods (such as LoRA) \cite{peft, lora} freeze most of the pre-trained model parameters and only fine-tune a small number of extra parameters. This makes fine-tuning both computation and storage efficient.

\noindent \textbf{Deepspeed:} Deepspeed \cite{deepspeed}, is a deep learning optimization toolkit that enables optimized training and inference of large deep learning models. 
We specifically use Zero Redundancy Optimizer (ZeRO)~\cite{zero} on Deepspeed for parallel training on 2 80G GPU cards. 
ZeRO reduces the memory requirement, speeds up training and allows scaling of the model size with the number of devices efficiently.

\subsection{Knowledge Fine-tuning}
We define Knowledge Fine-tuning (KFT) as fine-tuning a pre-trained model on a specialized dataset for improved understanding of context, grammar and terminologies used in the particular field. KFT imparts new knowledge into the pre-trained LLM.  A benefit of KFT is that it does not require expensive, labeled data. For this experiment we use National Archives (NARA) documents obtained via querying the records using their API. 

\noindent \textbf{NARA dataset:} We queried the NARA website and downloaded all of the text-based data available.  Many of the documents are only stored in \texttt{.pdf} or \texttt{.jpg} format, and so we had to perform optical character recognition (OCR) via PyPDF2 on those documents.  As a result, we had to clean the data: (i) we pick only the subset of documents that have more than 70\% characters that are alpha-numeric, (ii) next, from these, we pick only those files that have more than 70\% dictionary words. In both cases, the 70\% threshold is arbitrary but anecdotally led to human-readable texts. We then split this set into training and test sets. Our training set contains 6,263 articles and test set contains 1,566 articles. With the tokenization (sequences of length 512 tokens with overlap of 64 tokens), this gave us 94,441 training sequences and 23,442 test sequences. Our goal in using this dataset is to fine-tune a foundation model on the articles in National Archives so that the model learns the specific language of the government documents. 
The NARA archives were only queryable by API starting in 2023. Therefore, the Llama model was trained even before the NARA documents were queryable. That in addition to the fact that most NARA documents required further OCR makes us believe that this set of documents was not used for pre-training of the foundation models used in our experiment. That is one of the reasons we picked Llama model for this study.

\noindent \textbf{Training methodology:}
We leverage Hugging Face transformers API for all of our training\footnote{https://huggingface.co/docs/transformers}.
We leverage the Causal Language Modeling (Causal LM) framework %\footnote{\url{https://huggingface.co/docs/transformers/en/tasks/language\_modeling}} 
(\texttt{AutoModelForCausalLM}) for KFT. We used the Llama (7B) model as our foundation model and fine-tuned it via Causal LM on the above NARA dataset. 

\subsection{Format Fine-tuning}
We define format fine-tuning (FFT) as fine-tuning a pre-trained model for a specific task that requires strict output structure such as document summarization or Question/Answering with attribution. In this experiment we pick two news datasets described below and fine-tune a foundation model for document summarization. 
While training, we input the news articles for data and ground truth summaries as labels. 

\noindent \textbf{Kaggle news summary dataset:} 
We leveraged news dataset from Kaggle\footnote{https://www.kaggle.com/datasets/sunnysai12345/news-summary}. Reference summaries are generated by the inshorts\footnote{https://inshorts.com} app. 

\noindent \textbf{Newsroom dataset:}
We also leveraged a subset of the newsroom dataset\footnote{https://lil.nlp.cornell.edu/newsroom/index.html} for our study. Cornell newsroom is a large dataset 1.3 million articles and summaries written by authors and editors in the newsrooms of 38 major publications. The reference/ground truth summaries are obtained from search and social metadata between 1998 and 2017 and use a variety of summarization strategies combining extraction and abstraction. We picked a subset of this dataset with about 300K articles for our study. %We will release this dataset. 

\noindent \textbf{Training methodology:}
We leverage the Sequence to Sequence Language Modeling (\texttt{AutoModelForSeq2SeqLM} from Hugging Face transformers API). We used the Google T5 Small model~\cite{T5} for this study. 
We used the T5 model here as opposed to the Llama model as the encode-decoder structure of T5 is more amenable to sequence to sequence modeling. 

We include additional details on choice of training parameters in Appendix \ref{ssec:training} for both KFT and FFT described above.
\section{Comparison Metrics}
\label{sec:metrics}
We generate report summaries using foundation as well as fine-tuned models (details are in Appendix Subsection~\ref{ssec:generate}). We need to evaluate the generated summaries to ensure they represent the original articles accurately. 
One way to achieve this would be to have humans review the summaries and evaluate them. However, this approach is very labor-intensive and presents the challenge of recruiting human evaluators (expert or novice). Therefore in this paper we look for metrics for automatic evaluation of the generated summaries. We propose to use multiple metrics in this paper and describe them below. We perform manual inspection on a few articles to benchmark these metrics.

\subsection{Classical Metrics for Evaluating Text Summarization}
Numerous classical metrics are available for assessing text summarization, we use a subset of these in our study: the well known ROUGE~\cite{rouge}, BLEU~\cite{bleu_paper}, METEOR~\cite{meteor} and BERTScore~\cite{bertscore_paper}. Having a ground-truth summary is either explicitly required or heavily encouraged to use these metrics. Therefore, they are more suitable in the cases where such a reference summary is available. We use them in our FFT evaluation on news datasets due to availability of reference summaries. 

\subsection{Number of Valid Summaries}
We notice in some cases that LLMs can produce invalid summaries. We found this a challenge with the Llama 7B model on NARA dataset - one of the reasons for this could be the errors and noise in the NARA text due to OCR.  We use heuristics to identify invalid summaries. 

Some of the heuristics we include are as follows:
(i) Python function signature seen in a summary. 
(ii) The word ``summary'' occurs thrice or more in a summary.
(iii) The phrase "adobe reader" appears in a summary.
(iv) There is no letter from the English alphabet in a summary.
(v) Equal to notation (``='') occurs twice or more in a summary.

These heuristics are based on our manual observation of some of the generated summaries. There will be false positives and false negatives with this approach, but quantifying these at scale is challenging as it requires humans to review all of the generated summaries. However, the effectiveness of this approach in filtering out invalid summaries is demonstrated by the fact that the similarity metrics improve. We also perform a small human evaluation on a subset of summaries. We will provide more details in Section~\ref{sec:evaluate}. We show examples of invalid summaries in Appendix Subsection~\ref{ssec:appendix_invalid_eg}.

\subsection{Topic Similarity}
We also leverage a metric based on topic modeling in our study and refer to it as Topic Similarity (TS). A similar measure has been used in \cite{topic_similarity_related}. We use TS in our study as it is (i) interpretable and (ii) more suitable in the case where ground truth summaries may not be available. Computing TS involves the following steps:
\begin{itemize}
    \item Step 1: Represent a document by its topics via leveraging a topic modeling approach like the Latent Dirichlet Allocation (LDA)~\cite{lda}.
    \item Step 2: Compute the cosine similarity of these representations.
\end{itemize}

Latent Dirichlet Allocation (LDA)~\cite{lda} is a widely used topic modeling approach that helps discover underlying topics and themes within documents. 
LDA outputs the representation of a document as a vector of probabilities. The length of the vector is equal to $K$ which is the number of topics we input. Elements of the vector are probabilities, each corresponding to the document belonging to or containing the particular topic. 
%each of the $K$ topics. 
Note that each document can have multiple topics. 
We show examples of topic modeling outputs in Appendix Subsection~\ref{ssec:appendix_topic_modeling}. 
Topic Similarity simply computes the cosine similarity of these vectors output by LDA. 

Concretely, LDA algorithm will learn and output a vector $\boldsymbol{v_i}$ of length $K$, the number of topics, per document, where $i$ denotes the document ID. The document could be an article itself or a generated summary. Suppose we have article $a$ and corresponding generated summary $s$. LDA outputs two vectors for these documents, $\boldsymbol{v_a}$ and $\boldsymbol{v_s}$. To compute the relevance between the article and the summary, we compute the cosine similarity of their LDA outputs as follows where $\lvert\lvert \cdot \rvert \rvert$ is the length of the vector:
\begin{equation}
    TS(v_{a},v_{s}) = \dfrac{\boldsymbol{v_a} \cdot \boldsymbol{v_s}}{\lvert\lvert \boldsymbol{v_a} \rvert\rvert \ \lvert\lvert\boldsymbol{v_s}\rvert\rvert} 
\end{equation}

Given that we are performing a dimension reduction on all of the documents (articles and the summaries) by representing them in the form of vectors (output of LDA), before computing the similarity, the intuition is that the size of the document matters less. Therefore TS is more suitable when directly comparing closeness of a generated summary with the original article (as opposed to comparing the generated summary with a reference summary). 

\subsection{AlignScore}
AlignScore is a metric proposed in~\cite{alignscore_paper}. It is a tool for automatic evaluation of factual consistency. AlignScore showed promising results as per other works like~\cite{tofu_eval_paper}. It is based on a general function of information alignment between two text pieces.
They develop a training framework of alignment function and the training data comes from different NLP tasks. In certain instances, the reference text may be longer than the test text (for example, an article compared to its summary). In such cases, AlignScore will break the larger text into chunks, calculate the alignment between each chunk and the test text, and then use the highest alignment score from each chunk to compute an average for the final metric.

\subsection{Human Inspection}
Since human evaluation over entire datasets used is very challenging, we perform human inspection over small subsets of data in two important cases: (i) heuristices to detect invalid summaries: we read 100 NARA article summaries to determine if the summary is valid. Based on these 100 articles we provide false postitive rate and false negative rate of our heuristic based invalid summary detection, and (ii) correlation with automated metrics: in case of 50 NARA articles, we determine the usefulness of the summary through human evaluation and compare with the automated metrics.
\section{Evaluating Summaries}
\label{sec:evaluate}

\subsection{Knowledge Fine-tuning}
We study Knowledge Fine-tuning using the Llama 7B model on the NARA documents as explained in Section~\ref{sec:tune}. Our dataset has 6263 training documents and 1566 test documents. Our summary generation produces a slightly different summary each time it is invoked, so we generate three summaries from each article and look at the metrics across these three different runs in this subsection. Recall that (i) invalid summaries were a challenge in this setting and (ii) this set of documents has no ground truth labels. Therefore we evaluate the number of invalid summaries, Topic Similarity and AlignScore metrics.

\begin{figure}[ht]
\vspace{-2mm}
\begin{center}
\includegraphics[height=2in, width=3in]{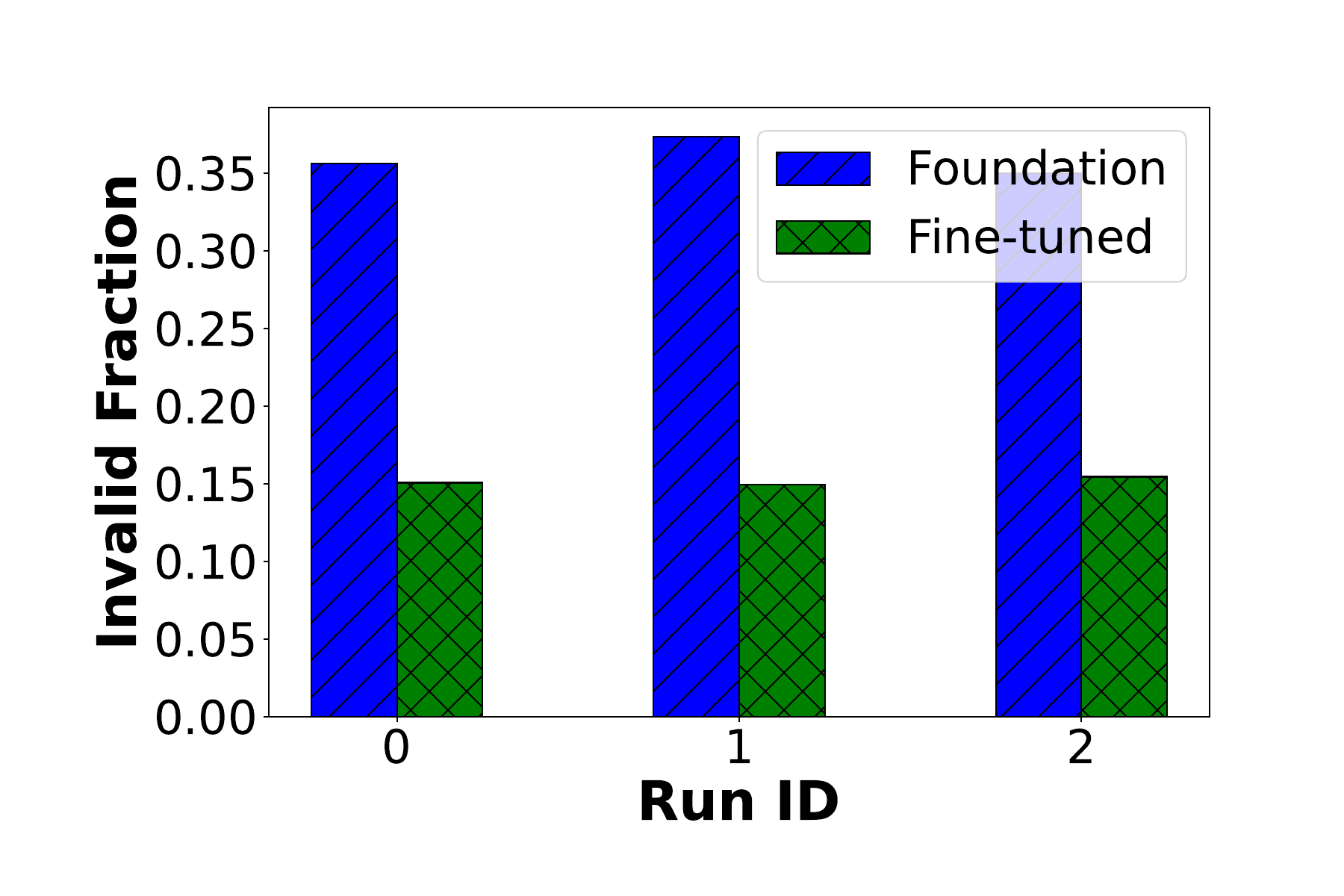}
\caption{Proportion of Invalid Summaries (Total = 1566).}
\label{fig:invalid}
\end{center}
%\Description{Proportion of invalid summaries.}
\vspace{-2mm}
\end{figure}

\noindent \textbf{Invalid Summaries:}
We pick invalid summaries based on the heuristics described in Section~\ref{sec:metrics}. Figure~\ref{fig:invalid} shows the findings. We see that the average proportion of invalid summaries goes down from about $36\%$ to $15\%$ with the knowledge fine tuning described in Section~\ref{sec:tune}. Possible explanations for this are as follows: first, the model learns to deal with noise within the NARA documents, and second these documents are very text-based as opposed to code or software based. This may encourage the model to produce natural language text as opposed to invalid summaries containing content like code snippets.

\begin{table*} [!h]
\begin{center}
\begin{tabular}{| c | c | c | c |}
\hline
 Mean $\pm$Stddev of TS w/ invalid & Run 0 & Run 1 & Run 2 \\ 
 \hline
 \hline
 Article - Foundation Summary & 0.69$\pm$ 0.40 & 0.70 $\pm$ 0.36 & 0.69 $\pm$ 0.39 \\ 
 \hline 
 Article - Fine-tuned Summary & 0.79$\pm$0.34 & 0.77$\pm$0.33 & 0.78$\pm$0.34 \\
 \hline
 Article - Random Summary &  0.32$\pm$0.40 & 0.33$\pm$0.38 & 0.30$\pm$0.37\\
 \hline 
\end{tabular}
\end{center}
\caption{Topic Similarity of Generated Summaries with the Original Article: Including the Invalid Summaries.}
\label{tab:ts_w_invalid}
\vspace{-2mm}
\end{table*}

\begin{table*} [!h]
\begin{center}
\begin{tabular}{| c | c | c | c |}
\hline
 Mean$\pm$Stddev of TS (w/o invalid) & Run 0 & Run 1 & Run 2 \\ 
 \hline
 \hline
 Article - Foundation Summary & 0.90$\pm$0.21 & 0.88$\pm$0.25 & 0.89$\pm$0.23 \\ 
 \hline 
 Article - Fine-tuned Summary & 0.84$\pm$0.26 & 0.82$\pm$0.29 & 0.84$\pm$0.27 \\
 \hline
 Article - Random Summary &  0.40$\pm$0.36 & 0.34$\pm$0.37 & 0.36$\pm$0.36\\
 \hline 
\end{tabular}
\end{center}
\caption{Topic Similarity of Generated Summaries with the Original Article: Excluding the Invalid Summaries.}
\label{tab:ts_wo_invalid}
\vspace{-2mm}
\end{table*}

Next, we manually inspect 100 summaries generated by the foundation model. % (during Run ID 0). 
39 articles were determined to be invalid. We find that our invalid summary detector was accurate: False Positives = 5, False Negatives = 7, True Positives = 32, True Negatives = 56. False Positive Rate = 0.08, False Negative Rate = 0.18. Note that this is only an initial filtering step, and a summary deemed valid does not automatically indicate that it is of high quality. Also, there is subjectivity associated with human inspection because some summaries have combination of valid and invalid content. 

\noindent \textbf{Topic Similarity:}
For this experiment, we compute LDA with 8 topics and 30 iterations. While ensuring that the topics are distinct, having $K=8$ still allows for coherent clustering of related words within each topic. Table~\ref{tab:ts_w_invalid}, shows the Topic Similarity before and Table~\ref{tab:ts_wo_invalid}, shows the same after filtering out invalid summaries. We compute TS of the article with a randomly picked summary (which may not belong to the current article) as a baseline for comparison purposes. Here are our observations:
\begin{itemize}
    \item Invalid summary detection: is demonstrated to be effective, since mean TS of foundation model summaries as well as fine-tuned model summaries improves and standard deviation decreases from Table~\ref{tab:ts_w_invalid} to Table~\ref{tab:ts_wo_invalid}. Average of Mean TS across the 3 runs improved by about 28\% for the foundation model and by 7\% for fine-tuned model. Average of Stddev of TS across 3 runs also improved, it decreased by about 40\% for foundation model and by about 19\% for the fine-tuned model. 
    \item Overall, the fine-tuned model outperforms the foundation model (in terms of TS metric) when we do not filter out the invalid summaries. Once invalid summaries are filtered out, foundation model outperforms the fine-tuned model. Therefore, under this setting the main utililty of KFT is to reduce the number of invalid summaries. 
    \item As expected, mean TS between an article and random summary is low and the stddev is high (compared to both the foundation model summaries and fine-tuned model summaries), this serves more as a sanity check.
    \item Even the best scores have considerably high standard deviations ($>0.2$), which indicates inconsistency of summary quality with these models. We experimented with Llama 7B as it was trained before NARA dataset was available and so KFT study was more interesting. 
\end{itemize}

\begin{table*} [!h]
\begin{center}
\begin{tabular}{| c | c | c | c |}
\hline
 Mean of AlignScore w/ invalid & Run 0 & Run 1 & Run 2 \\ 
 \hline
 \hline
 Article - Foundation Summary & 0.49 & 0.49 & 0.48 \\ 
 \hline 
 Article - Fine-tuned Summary & 0.45 & 0.46 & 0.46 \\
 \hline
\end{tabular}
\end{center}
\caption{AlignScore of Generated Summaries with the Original Article: Including the Invalid Summaries.}
\label{tab:alignscore_w_invalid}
\vspace{-2mm}
\end{table*}

\begin{table*} [!h]
\begin{center}
\begin{tabular}{| c | c | c | c |}
\hline
Mean of AlignScore  (w/o invalid) & Run 0 & Run 1 & Run 2 \\ 
 \hline
 \hline
 Article - Foundation Summary & 0.49 & 0.49 & 0.48 \\ 
 \hline 
 Article - Fine-tuned Summary & 0.45 & 0.46 & 0.46 \\
 \hline
\end{tabular}
\end{center}
\caption{AlignScore of Generated Summaries with the Original Article: Excluding the Invalid Summaries.}
\label{tab:alignscore_wo_invalid}
\vspace{-2mm}
\end{table*}

\noindent \textbf{AlignScore:} We summarize the results in Table~\ref{tab:alignscore_w_invalid} and Table~\ref{tab:alignscore_wo_invalid}. The results agree with TS metric after invalid summaries are removed. However we find that AlignScore does not improve as much when invalid summaries are removed. We examine this further in Appendix Subsection~\ref{ssec:alignscore}. We find that AlignScore worked well for cases where summaries maybe on topic but not usable, but it also showed high score on an average even for invalid summaries. This could be because of how chunking and aggregating over chunks is done - using maximum value of similarity. Further, for Run ID 0, we computed mean AlignScore of documents with randomly picked summary, this was 0.36 - which is lower than the scores for the correct summaries (in the tables) but difference is smaller than the case of TS above. Therefore we find that always using multiple metrics is necessary to get a more accurate understanding of the performance.

\noindent \textbf{Human Inspection:} We manually inspect 50 valid summaries from foundation Llama 7B model. We use two types of categorization: (i) binary: useful, not useful, and (ii) three way classification: 0, 1, 2, with 0 least being least useful and 2 being most useful. Note that the classification can be subjective but gives a reasonable estimate of how human evaluation might correlate with the automated metrics. We outline the result in Table~\ref{tab:human_eval2} and Table~\ref{tab:human_eval3}. Notice that although small, this evaluation shows that both AlignScore and Topic Similarity are higher for more useful summaries and lower for less useful summaries.

\begin{table*} [!h]
%\begin{center}
\begin{minipage}{.5\linewidth}
\begin{center}
\begin{tabular}{| c | c | c | }
\hline
  & TS & AlignScore \\ 
 \hline
 \hline
 Not Useful & 0.81 & 0.40  \\
 \hline 
 Useful & 0.99 & 0.54  \\ 
 \hline
\end{tabular}
%\end{center}
\caption{Categorization I.}
\label{tab:human_eval2}
%\begin{center}
\end{center}
\end{minipage}%
\begin{minipage}{.5\linewidth}
\begin{center}
\begin{tabular}{| c | c | c |}
\hline
  & TS & AlignScore  \\ 
 \hline
 \hline
 Useful Rated: 0 & 0.81 & 0.40  \\ 
 \hline 
 Useful Rated: 1 & 0.97 & 0.47  \\
 \hline
 Useful Rated: 2 & 0.98 & 0.56  \\
 \hline
\end{tabular}
%\end{center}
\caption{Categorization II.}
\label{tab:human_eval3}
\end{center}
\end{minipage}%
\vspace{-2mm}
\end{table*}

\subsection{Format Fine-tuning}
\label{ssec:fft_news_eval}
As described in Section~\ref{sec:tune}, we fine-tune the Google T5 model for the FFT experiments. Since the T5 model generates the same summary for an article every time it is invoked, we summarize every article only once using the T5 model. We evaluate the model on two different news datasets. % as we describe below.
\subsubsection{Kaggle News Summary Experiments}
\label{sssec:fft_news_eval_kaggle}
Here we report our findings from the Kaggle News Summary dataset. Our training set has 3553 articles and test set has 903 articles.

%\noindent \textbf{Classical Metrics:}
In Table \ref{tab:rouge_kaggle} we compute the classical metrics as explained in Section~\ref{sec:metrics}. For the Kaggle dataset, we find that with FFT, the model helps generate summaries that are closer to the ground truth summaries. We see 25\% improvement in ROUGE-1, 50\% in ROUGE-2, 27\% in ROUGE-L, 138\% in BLEU, 54\% in METEOR and 2\% in BERTScore metric due to fine-tuning T5 model on this dataset.

\begin{table*} [!h]
\begin{center}
\begin{tabular}{| c | c | c | c | c | c | c | }
\hline
  & ROUGE-1 & ROUGE-2 & ROUGE-L & BLEU & METEOR & BERTScore \\ 
 \hline
 \hline
 Foundation & 0.36 & 0.16 & 0.26 & 0.08 & 0.24 & 0.87 \\ 
 \hline 
 Fine-tuned & 0.45 & 0.24 & 0.33  & 0.19 & 0.37 & 0.89 \\
 \hline
\end{tabular}
\end{center}
\caption{Classical metrics on Kaggle dataset T5 model summaries.}
\label{tab:rouge_kaggle}
\vspace{-4mm}
\end{table*}

\begin{table*} [!h]
\begin{center}
\begin{tabular}{| c | c | c | c | c | c | c |}
\hline
  & ROUGE-1 & ROUGE-2 & ROUGE-L & BLEU & METEOR & BERTScore \\ 
 \hline
 \hline
 Foundation & 0.24 & 0.09 & 0.18 & 0.07 & 0.24 & 0.85 \\ 
 \hline 
 Fine-tuned & 0.36 & 0.24 & 0.31  & 0.20 & 0.39 & 0.88 \\
 \hline
\end{tabular}
\end{center}
\caption{Classical metrics on Newsroom dataset T5 model summaries.}
\label{tab:rouge_newsroom}
\vspace{-4mm}
\end{table*}

In Table \ref{tab:ts_kaggle}, we also compute TS on the same set of generated summaries. Here, we compare the generated summaries against the original article as the classical metrics are more suitable and established for comparing generated summaries against the reference summaries. 
We ran LDA topic modeling with 30 topics and 30 iterations for this study. Again, we found this number to be a good balance for the number of topics in the news datasets. We find that TS metric also confirms that fine-tuning helps generate better summaries. We find that fine-tuning generates summaries equivalent to the reference (ground truth) summaries, and improves mean TS metric by about 10\% over the foundation model summaries. It may be worth noting that this set of summaries was generated by another AI technology and thus the result is not surprising. We further see that the standard deviation of TS is also improved due to the fine-tuning (by about 20\% over foundation model summaries).
\begin{table} [!h]
\begin{center}
\begin{tabular}{| c | c |}
\hline
 & Mean$\pm$Stddev of TS   \\ 
 \hline
 \hline
 Article - Reference & 0.85$\pm$0.25  \\ 
 \hline 
 Article - Foundation & 0.78$\pm$0.30 \\ 
 \hline 
 Article - Fine-tuned & 0.86$\pm$0.24  \\
 \hline
 Article - Random &  0.48$\pm$0.39 \\
 \hline 
\end{tabular}
\end{center}
\caption{TS metric on Kaggle dataset T5 model summaries.}
\label{tab:ts_kaggle}
\vspace{-4mm}
\end{table}

\subsection{Newsroom Experiments} 
\label{ssec:newsroom} 
We conducted similar experiments on our Newsroom dataset with $77,567$ training articles and $19,402$ test articles. From Table~\ref{tab:rouge_newsroom} and Table~\ref{tab:ts_newsroom}, we see the trends are similar to the results from Kaggle news summary dataset above. In addition, from Table~\ref{tab:ts_newsroom}, we find that both foundation model and fine-tuned model generate summaries closer to the article than the reference summary - this is possible based on how the reference (ground truth) summaries are generated.

\begin{table} [!h]
\begin{center}
\begin{tabular}{| c | c |}
\hline
 & Mean$\pm$Stddev of TS    \\ 
 \hline
 \hline
 Article - Reference & 0.48$\pm$0.39  \\ 
 \hline 
 Article - Foundation & 0.58$\pm$0.38 \\ 
 \hline 
 Article - Fine-tuned & 0.60$\pm$0.38  \\
 \hline
 Article - Random &  0.17$\pm$0.28 \\
 \hline 
\end{tabular}
\end{center}
\caption{TS metric on Newsroom dataset T5 model summaries.}
\label{tab:ts_newsroom}
\vspace{-4mm}
\end{table}

\section{Related Work}
\label{sec:related} 
We divide the related work into the following two categories: (i) Fine-tuning LLMs, and (ii) Metrics for evaluating text summarization.

\noindent \textbf{Fine-tuning LLMs:} 
A similar study of utility of fine-tuning for the specific case of finance is conducted in \cite{finetuning_finance}, in comparison, our study focuses on general utility of fine-tuning and fine-tuning for government use cases. Authors in \cite{finetune_facts} fine-tune LLMs for factual correctness. Work in \cite{finetune_hallucination} studies the risk involved in introducing LLMs to new knowledge during fine-tuning, and find that this increases the model’s tendency to hallucinate and therefore suggest that fine-tuning may be more useful to teach the model to use its knowledge more efficiently versus introducing new knowledge. There is also a line of work that studies methods like regularization and parameter selection in fine-tuning to avoid over-fitting of the models \cite{finetune_regularize2, finetune_regularize3, finetune_regularize1}. These works are complementary to our work and study different aspects of fine-tuning. 

\noindent \textbf{Metrics for evaluating text summarization:} We have leveraged many of the classical metrics for evaluating automatic text summarization in Section~\ref{sec:metrics}. There has been research to improve on these metrics. For example, NIST~\cite{nist} modifies BLEU to weigh n-grams differently and use different brevity penalty. Moverscore~\cite{moverscore} improves upon BERTScore via adding soft alignments. Automatic metrics are combined with human evaluation in \cite{metric1, metric2}. Other research in the area includes G-Eval~\cite{geval} which uses LLMs with Chain of Thoughts (CoT) to input the original article with the reference text to evaluate the quality of generated summary. While these are related to our work, our work mainly focuses on selecting explainable metrics that are suitable for evaluating text summarization tasks in the wild. We report our results using multiple metrics for improved confidence in our findings.

\section{Conclusion and Future Work}
\label{sec:conclude}
In this paper, we study the specific problem of report summarization on datasets of interest to the government. First, we collect and curate a dataset of National Archives data for the study. Second, we pick multiple metrics and design heuristics we could use in our setting where ground truth summaries may not always be available. Finally, using these we see that (i) it is feasible to leverage modest sized LLMs (Llama 7B and T5 Small) for reasonable quality summaries, (ii) we can fine-tune these LLMs efficiently using techniques like PEFT LoRA and Deepspeed ZeRO for improved summarization. Specifically results with fine-tuning were interesting, a) Knowledge Fine Tuning (KFT) on National Archives helped reduce the invalid summaries generated by Llama 7B, however, after eliminating the invalid summaries, a comparison of the valid summaries alone showed that the summaries generated by the fine-tuned model did not exhibit superior quality to those from the foundation model, b) Format Fine Tuning (FFT) T5 Small on news datasets consistently improved the summary quality unlike the case of KFT above. 

However a lot of open questions remain. Some of the important ones include: how does varying PEFT parameters impact the fine-tuning? what is the reason for invalid summaries, how do they manifest with different datasets, models and different prompting strategies, how to improve the invalid summary checker? There are some practical benefits of fine-tuning over few shot prompting (more accurate on larger models, requires prompt engineering) or retrieval augmented generation (requires long context window), however, how would these approaches perform in comparison with fine-tuning approaches for the use-cases we consider? 

\section*{Acknowledgements}
%This material is based upon work funded and supported by the Department of Defense under Contract No. FA8702-15-D-0002 with Carnegie Mellon University for the operation of the Software Engineering Institute, a federally funded research and development center. The view, opinions, and/or findings contained in this material are those of the author(s) and should not be construed as an official Government position, policy, or decision, unless designated by other documentation. This work is licensed under a Creative Commons Attribution-NonCommercial 4.0 International License.  
Copyright 2026 Carnegie Mellon University. 

This material is based upon work funded and supported by the Independent Agency,  under Air Force Contract No. FA8702-15-D-0002 with Carnegie Mellon University for the operation of the Software Engineering Institute, a federally funded research and development center sponsored by the United States Department of War. 

The opinions, findings, conclusions, and/or recommendations contained in this material are those of the author(s) and should not be construed as an official US Government position, policy, or decision, unless designated by other documentation. 

References herein to any specific entity, product, process, or service by trade name, trademark, manufacturer, or otherwise, does not necessarily constitute or imply its endorsement, recommendation, or favoring by Carnegie Mellon University or its Software Engineering Institute nor of Carnegie Mellon University - Software Engineering Institute by any such named or represented entity.
 
[DISTRIBUTION STATEMENT A] This material has been approved for public release and unlimited distribution.  Please see Copyright notice for non-US Government use and distribution.
 
This work is licensed under a Creative Commons Attribution-NonCommercial 4.0 International License (https://creativecommons.org/licenses/by-nc/4.0/).  Requests for permission for non-licensed uses should be directed to the Software Engineering Institute at permission@sei.cmu.edu.
 
DM25-0140

\bibliography{metrics-bib}

@inproceedings{lda,
author = {Blei, David and Ng, Andrew and Jordan, Michael},
year = {2001},
month = {01},
pages = {601-608},
title = {Latent Dirichlet Allocation},
volume = {3},
journal = {The Journal of Machine Learning Research}
}

@inproceedings{rouge,
author = {Lin, Chin-Yew},
title = {ROUGE: a Package for Automatic Evaluation of Summaries},
booktitle = {Workshop on Text Summarization Branches Out, Post-Conference Workshop of ACL 2004, Barcelona, Spain},
year = {2004},
month = {July}
}

@article{t5,
  author  = {Colin Raffel and Noam Shazeer and Adam Roberts and Katherine Lee and Sharan Narang and Michael Matena and Yanqi Zhou and Wei Li and Peter J. Liu},
  title   = {Exploring the Limits of Transfer Learning with a Unified Text-to-Text Transformer},
  journal = {Journal of Machine Learning Research},
  year    = {2020},
  volume  = {21},
  number  = {140},
  pages   = {1-67},
  url     = {http://jmlr.org/papers/v21/20-074.html}
}

@Misc{peft,
  title =        {PEFT: State-of-the-art Parameter-Efficient Fine-Tuning methods},
  author =       {Sourab Mangrulkar and Sylvain Gugger and Lysandre Debut and Younes Belkada and Sayak Paul and Benjamin Bossan},
  howpublished = {\url{https://github.com/huggingface/peft}},
  year =         {2022}
}

@misc{deepspeed,
  title = {Deepspeed},
  author = {Microsoft},
  howpublished = {\url{https://github.com/microsoft/DeepSpeed}},
  year = {2023}
}

@article{zero,
  author       = {Samyam Rajbhandari and
                  Jeff Rasley and
                  Olatunji Ruwase and
                  Yuxiong He},
  title        = {ZeRO: Memory Optimization Towards Training {A} Trillion Parameter
                  Models},
  journal      = {CoRR},
  volume       = {abs/1910.02054},
  year         = {2019},
  url          = {http://arxiv.org/abs/1910.02054},
  eprinttype    = {arXiv},
  eprint       = {1910.02054},
  bibsource    = {dblp computer science bibliography, https://dblp.org}
}

@article{lora,
  author       = {Edward J. Hu and
                  Yelong Shen and
                  Phillip Wallis and
                  Zeyuan Allen{-}Zhu and
                  Yuanzhi Li and
                  Shean Wang and
                  Weizhu Chen},
  title        = {LoRA: Low-Rank Adaptation of Large Language Models},
  journal      = {CoRR},
  volume       = {abs/2106.09685},
  year         = {2021},
  url          = {https://arxiv.org/abs/2106.09685},
  eprinttype    = {arXiv},
  eprint       = {2106.09685},
  bibsource    = {dblp computer science bibliography, https://dblp.org}
}

@INPROCEEDINGS{bleu_paper,
    author = {Kishore Papineni and Salim Roukos and Todd Ward and Wei-jing Zhu},
    title = {BLEU: a Method for Automatic Evaluation of Machine Translation},
    booktitle = {},
    year = {2002},
    pages = {311--318}
}

@article{bertscore_paper,
  author       = {Tianyi Zhang and
                  Varsha Kishore and
                  Felix Wu and
                  Kilian Q. Weinberger and
                  Yoav Artzi},
  title        = {BERTScore: Evaluating Text Generation with {BERT}},
  journal      = {CoRR},
  volume       = {abs/1904.09675},
  year         = {2019},
  url          = {http://arxiv.org/abs/1904.09675},
  eprinttype    = {arXiv},
  eprint       = {1904.09675},
  bibsource    = {dblp computer science bibliography, https://dblp.org}
}

@inproceedings{meteor,
    title = "{METEOR}: An Automatic Metric for {MT} Evaluation with Improved Correlation with Human Judgments",
    author = "Banerjee, Satanjeev  and
      Lavie, Alon",
    editor = "Goldstein, Jade  and
      Lavie, Alon  and
      Lin, Chin-Yew  and
      Voss, Clare",
    booktitle = "Proceedings of the {ACL} Workshop on Intrinsic and Extrinsic Evaluation Measures for Machine Translation and/or Summarization",
    month = jun,
    year = "2005",
    address = "Ann Arbor, Michigan",
    publisher = "Association for Computational Linguistics",
    url = "https://aclanthology.org/W05-0909",
    pages = "65--72",
}

@misc{geval,
      title={G-Eval: NLG Evaluation using GPT-4 with Better Human Alignment}, 
      author={Yang Liu and Dan Iter and Yichong Xu and Shuohang Wang and Ruochen Xu and Chenguang Zhu},
      year={2023},
      eprint={2303.16634},
      archivePrefix={arXiv},
      primaryClass={cs.CL}
}

@misc{moverscore,
      title={MoverScore: Text Generation Evaluating with Contextualized Embeddings and Earth Mover Distance}, 
      author={Wei Zhao and Maxime Peyrard and Fei Liu and Yang Gao and Christian M. Meyer and Steffen Eger},
      year={2019},
      eprint={1909.02622},
      archivePrefix={arXiv},
      primaryClass={cs.CL}
}

@article{nist,
author = {Doddington, George},
year = {2002},
month = {01},
pages = {138-145},
title = {Automatic evaluation of machine translation quality using n-gram co-occurrence statistics},
doi = {10.3115/1289189.1289273}
}

@inproceedings{metric1,
    title = "The price of debiasing automatic metrics in natural language evalaution",
    author = "Chaganty, Arun  and
      Mussmann, Stephen  and
      Liang, Percy",
    editor = "Gurevych, Iryna  and
      Miyao, Yusuke",
    booktitle = "Proceedings of the 56th Annual Meeting of the Association for Computational Linguistics (Volume 1: Long Papers)",
    month = jul,
    year = "2018",
    address = "Melbourne, Australia",
    publisher = "Association for Computational Linguistics",
    url = "https://aclanthology.org/P18-1060",
    doi = "10.18653/v1/P18-1060",
    pages = "643--653"
}

@inproceedings{metric2,
    title = "Unifying Human and Statistical Evaluation for Natural Language Generation",
    author = "Hashimoto, Tatsunori B.  and
      Zhang, Hugh  and
      Liang, Percy",
    editor = "Burstein, Jill  and
      Doran, Christy  and
      Solorio, Thamar",
    booktitle = "Proceedings of the 2019 Conference of the North {A}merican Chapter of the Association for Computational Linguistics: Human Language Technologies, Volume 1 (Long and Short Papers)",
    month = jun,
    year = "2019",
    address = "Minneapolis, Minnesota",
    publisher = "Association for Computational Linguistics",
    url = "https://aclanthology.org/N19-1169",
    doi = "10.18653/v1/N19-1169",
    pages = "1689--1701"
}

@misc{finetune_hallucination,
      title={Does Fine-Tuning LLMs on New Knowledge Encourage Hallucinations?}, 
      author={Zorik Gekhman and Gal Yona and Roee Aharoni and Matan Eyal and Amir Feder and Roi Reichart and Jonathan Herzig},
      year={2024},
      eprint={2405.05904},
      archivePrefix={arXiv},
      primaryClass={cs.CL}
}

@misc{finetune_regularize1,
      title={Fine-Tuning Pre-Trained Language Models Effectively by Optimizing Subnetworks Adaptively}, 
      author={Haojie Zhang and Ge Li and Jia Li and Zhongjin Zhang and Yuqi Zhu and Zhi Jin},
      year={2022},
      eprint={2211.01642},
      archivePrefix={arXiv},
      primaryClass={cs.CL}
}

@article{finetune_regularize2,
  author       = {Cheolhyoung Lee and
                  Kyunghyun Cho and
                  Wanmo Kang},
  title        = {Mixout: Effective Regularization to Finetune Large-scale Pretrained
                  Language Models},
  journal      = {CoRR},
  volume       = {abs/1909.11299},
  year         = {2019},
  url          = {http://arxiv.org/abs/1909.11299},
  eprinttype    = {arXiv},
  eprint       = {1909.11299},
  bibsource    = {dblp computer science bibliography, https://dblp.org}
}

@article{finetune_regularize3,
  author       = {Runxin Xu and
                  Fuli Luo and
                  Zhiyuan Zhang and
                  Chuanqi Tan and
                  Baobao Chang and
                  Songfang Huang and
                  Fei Huang},
  title        = {Raise a Child in Large Language Model: Towards Effective and Generalizable
                  Fine-tuning},
  journal      = {CoRR},
  volume       = {abs/2109.05687},
  year         = {2021},
  url          = {https://arxiv.org/abs/2109.05687},
  eprinttype    = {arXiv},
  eprint       = {2109.05687},
  bibsource    = {dblp computer science bibliography, https://dblp.org}
}

@misc{fewshot,
      title={Language Models are Few-Shot Learners}, 
      author={Tom B. Brown and Benjamin Mann and Nick Ryder and Melanie Subbiah and Jared Kaplan and Prafulla Dhariwal and Arvind Neelakantan and Pranav Shyam and Girish Sastry and Amanda Askell and Sandhini Agarwal and Ariel Herbert-Voss and Gretchen Krueger and Tom Henighan and Rewon Child and Aditya Ramesh and Daniel M. Ziegler and Jeffrey Wu and Clemens Winter and Christopher Hesse and Mark Chen and Eric Sigler and Mateusz Litwin and Scott Gray and Benjamin Chess and Jack Clark and Christopher Berner and Sam McCandlish and Alec Radford and Ilya Sutskever and Dario Amodei},
      year={2020},
      eprint={2005.14165},
      archivePrefix={arXiv},
      primaryClass={cs.CL}
}

@article{rag,
  author       = {Patrick S. H. Lewis and
                  Ethan Perez and
                  Aleksandra Piktus and
                  Fabio Petroni and
                  Vladimir Karpukhin and
                  Naman Goyal and
                  Heinrich K{\"{u}}ttler and
                  Mike Lewis and
                  Wen{-}tau Yih and
                  Tim Rockt{\"{a}}schel and
                  Sebastian Riedel and
                  Douwe Kiela},
  title        = {Retrieval-Augmented Generation for Knowledge-Intensive {NLP} Tasks},
  journal      = {CoRR},
  volume       = {abs/2005.11401},
  year         = {2020},
  url          = {https://arxiv.org/abs/2005.11401},
  eprinttype    = {arXiv},
  eprint       = {2005.11401},
  bibsource    = {dblp computer science bibliography, https://dblp.org}
}

@misc{finetuning_finance,
      title={Fine-tuning and Utilization Methods of Domain-specific LLMs}, 
      author={Cheonsu Jeong},
      year={2024},
      eprint={2401.02981},
      archivePrefix={arXiv},
      primaryClass={cs.CL}
}

@misc{finetune_facts,
      title={Fine-tuning Language Models for Factuality}, 
      author={Katherine Tian and Eric Mitchell and Huaxiu Yao and Christopher D. Manning and Chelsea Finn},
      year={2023},
      eprint={2311.08401},
      archivePrefix={arXiv},
      primaryClass={cs.CL}
}

@article{topic_similarity_related,
  title={Evaluation Measures for Text Summarization},
  author={Josef Steinberger and Karel Jezek},
  journal={Comput. Informatics},
  year={2012},
  volume={28},
  pages={251-275},
  url={https://api.semanticscholar.org/CorpusID:30451120}
}

@misc{llm_law1,
      title={SaulLM-7B: A pioneering Large Language Model for Law}, 
      author={Pierre Colombo and Telmo Pessoa Pires and Malik Boudiaf and Dominic Culver and Rui Melo and Caio Corro and Andre F. T. Martins and Fabrizio Esposito and Vera Lúcia Raposo and Sofia Morgado and Michael Desa},
      year={2024},
      eprint={2403.03883},
      archivePrefix={arXiv},
      primaryClass={cs.CL}
}

@misc{llm_medicine1,
      title={Domain-specific Continued Pretraining of Language Models for Capturing Long Context in Mental Health}, 
      author={Shaoxiong Ji and Tianlin Zhang and Kailai Yang and Sophia Ananiadou and Erik Cambria and Jörg Tiedemann},
      year={2023},
      eprint={2304.10447},
      archivePrefix={arXiv},
      primaryClass={cs.CL}
}

@inproceedings{parrot, author = {Bender, Emily M. and Gebru, Timnit and McMillan-Major, Angelina and Shmitchell, Shmargaret}, title = {On the Dangers of Stochastic Parrots: Can Language Models Be Too Big?}, year = {2021}, isbn = {9781450383097}, publisher = {Association for Computing Machinery}, address = {New York, NY, USA}, url = {https://doi.org/10.1145/3442188.3445922}, doi = {10.1145/3442188.3445922}, booktitle = {Proceedings of the 2021 ACM Conference on Fairness, Accountability, and Transparency}, pages = {610–623}, numpages = {14}, location = {Virtual Event, Canada}, series = {FAccT '21} }

@misc{tofu_eval_paper,
      title={TofuEval: Evaluating Hallucinations of LLMs on Topic-Focused Dialogue Summarization}, 
      author={Liyan Tang and Igor Shalyminov and Amy Wing-mei Wong and Jon Burnsky and Jake W. Vincent and Yu'an Yang and Siffi Singh and Song Feng and Hwanjun Song and Hang Su and Lijia Sun and Yi Zhang and Saab Mansour and Kathleen McKeown},
      year={2024},
      eprint={2402.13249},
      archivePrefix={arXiv},
      primaryClass={cs.CL},
      url={https://arxiv.org/abs/2402.13249}, 
}

@inproceedings{alignscore_paper,
    title = "{A}lign{S}core: Evaluating Factual Consistency with A Unified Alignment Function",
    author = "Zha, Yuheng  and
      Yang, Yichi  and
      Li, Ruichen  and
      Hu, Zhiting",
    editor = "Rogers, Anna  and
      Boyd-Graber, Jordan  and
      Okazaki, Naoaki",
    booktitle = "Proceedings of the 61st Annual Meeting of the Association for Computational Linguistics (Volume 1: Long Papers)",
    month = jul,
    year = "2023",
    address = "Toronto, Canada",
    publisher = "Association for Computational Linguistics",
    url = "https://aclanthology.org/2023.acl-long.634",
    doi = "10.18653/v1/2023.acl-long.634",
    pages = "11328--11348",
}

\appendix
\section{Appendix}
\label{sec:appendix}

\subsection{Details of Training Methodology}
\label{ssec:training}
We leverage Hugging Face transformers API for all of our training\footnote{https://huggingface.co/docs/transformers}.

\noindent \textbf{Knowledge Fine Tuning:}
We leverage the Causal Language Modeling (Causal LM) framework 
(\texttt{AutoModelForCausalLM}) for fine-tuning the Llama (7B) model on NARA dataset.
First, we performed tokenization using \texttt{LlamaTokenizer}. 
We take the entire article and tokenize it into sequences of size 512 tokens, with an overlap of 64 tokens. After tokenization, we use the input tokens also as labels. The model takes care of shifting the labels for Causal Language Modeling (Causal LM) (predicting next token). 
%~\cite{causalLM, auto_shift}. %Thus this is unsupervised training.
We trained the model for 3 epochs on 2 GPU cards using Deepspeed and PEFT with a batch size of 8 per card. We used Low-Rank Adaptation (LoRA)~\cite{lora} on PEFT with $r = 16$ and $alpha = 32$, which worked well in our experiments. 
Here, $r$ is the rank of the matrix used, and $alpha$ is a scaling parameter for the weights. In future work, we would like to evaluate the effect of varying these parameters. PEFT together with Deepspeed makes it feasible to train the model on our available hardware very efficiently. It took about 5 hours to fine-tune this model using 2 80G Nvidia A100 GPU cards on a single machine for our training. Once training is completed with PEFT, the smaller adapter model is saved and is merged back to the base model for future use. 

\noindent \textbf{Format Fine Tuning:}
We leverage the Sequence to Sequence Language Modeling (\texttt{AutoModelForSeq2SeqLM} from Hugging Face transformers API). We used the Google T5 Small model~\cite{T5} for this study. Note that this is a different task than KFT which only required a decoder model for next word prediction. We used the T5 model here as opposed to the Llama model as the encode-decoder transformer structure of T5 is more amenable to sequence to sequence modeling. 
%Article-to-summary FFT is arguably better suited an encoder-decoder model for this specialized task. 
Moreover, using the small T5 (60M parameters) model helps us with efficiency of training on our hardware.
For both the datasets we only picked articles that were less then 2048 tokens and whose summaries also were less than 512 tokens, to keep the training time efficient. This gave us 3553 training (903 test) articles in Kaggle news summary dataset and 77567 training (19402 test) articles in the Newsroom dataset. We first performed tokenization using the T5 small tokenizer from Hugging Face. We prefix the input with a prompt that says ``summarize: '' so that models knows it is a summarization task. The article is passed as input, and ground truth summary as the label. We used LoRA on PEFT as before with $r=16$ and $alpha=32$. We trained the Kaggle newsroom dataset for 30 epochs and newsroom for 10 epochs on one A100 80G GPU card with a batch size of 16. Fine-tuning on Kaggle news summary took about 30 minutes and fine-tuning on newsroom dataset took about 6 hours. After training, the PEFT adapter model was merged with the base model.

\begin{table*}[!ht]
\begin{center}
\begin{tabular}{| c | c |}
\hline
 Topic ID & Words \\ 
 \hline
 \hline
 Topic 0 & file document following already later using view together benefit within \\ 
 \hline 
 Topic 1 & dol osha gov violation safety department worker release labor news \\
 \hline
 Topic 2 & market agricultural food report rate committee product import percent meeting \\
 \hline 
 Topic 3 & pbgc plan benefit corporation senate washington timely performed investment employer  \\
 \hline 
 Topic 4 & corps peace book welcome development letter 2013 serve history programs \\
 \hline 
 Topic 5 & appellant board decision claim compensation injury owcp work office evidence \\
 \hline 
 Topic 6 & rule commission company wage mine employee hour exchange investigation labor  \\
 \hline 
 Topic 7 & attorney general justice united states crime department would people court \\
 \hline 
\end{tabular}
\end{center}
\caption{Example of Topics Generated from NARA documents and their summaries}
\label{tab:eg_topics}
\end{table*}

\begin{figure*}[ht]
\vspace{-2mm}
\begin{center}
\includegraphics[width=0.48\textwidth]{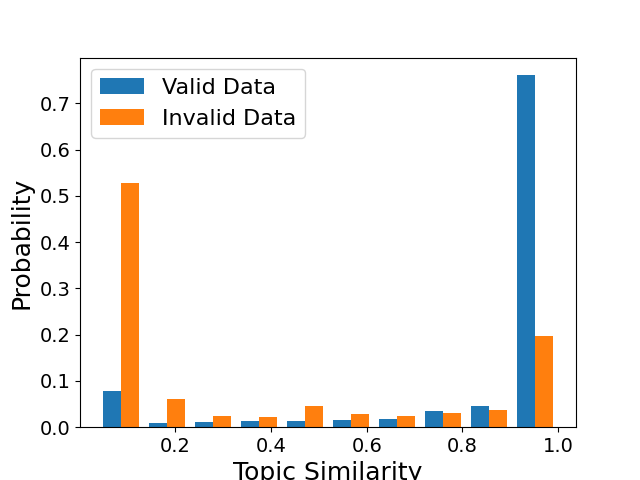}
\hfill
\includegraphics[width=0.48\textwidth]{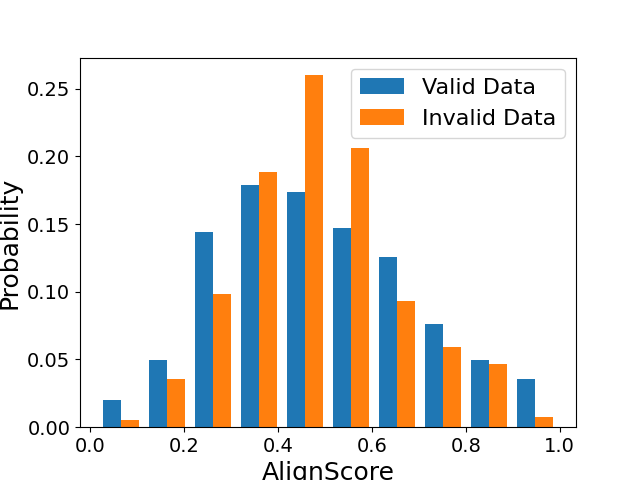}
\vspace{2mm}
\caption{Distribution of Topic Similarity and AlignScore for comparing articles with their corresponding summaries.}
\label{fig:as_ts_pdf}
\end{center}
%\vspace{-2mm}
\end{figure*}

\begin{table} 
%\vspace{-130mm}
\begin{center}
\begin{tabular}{| c | c |}
\hline
 Topic ID & Words \\ 
 \hline
 \hline
 Topic 0 & 1.55\% \\ 
 \hline 
 Topic 1 & 1.55\% \\
 \hline
 Topic 2 & 41.35\% \\
 \hline 
 Topic 3 & 1.55\%  \\
 \hline 
 Topic 4 & 1.55\% \\
 \hline 
 Topic 5 & 1.55\% \\
 \hline 
 Topic 6 & 1.55\%  \\
 \hline 
 Topic 7 & 49.33\% \\
 \hline 
\end{tabular}
\end{center}
\caption{Example of Topics Generated from a document.}
\label{tab:eg_doc_topics}
\vspace{-4mm}
\end{table}

\subsection{Generating Summaries}
\label{ssec:generate}
% Our next step is to perform text summarization using the foundation models as well as our fine-tuned models. 
% %We generate summaries using foundation models as well as our fine-tuned models. 
% As discussed in Section~\ref{sec:tune}, we study two types of fine-tuning methodologies: (i) Knowledge Fine-tuning (KFT), and (ii) Format Fine-tuning (FFT). 

% For Knowledge Fine-tuning, we use the NARA dataset with the Llama 7B model~\cite{llama}. 

\noindent \textbf{Llama 7B:} To generate summaries using Llama models (foundation and fine-tuned), we leverage the Langchain framework for summarization\footnote{https://python.langchain.com/docs/use\_cases/summarization}. 
%Langchain enables several applications that use LLMs. It supports running a sequence of function calls by combining LLMs with other functions. In case of summarization, it supports using different prompts with different LLMs and this helps with using the Llama model for text summarization. 
Langchain supports different summarization methodologies: (i) Stuff, (ii) Refine, and (iii) Map-reduce. Refine and Map-reduce methodologies can chunk up articles into smaller pieces and generate summaries and summary of these summaries iteratively. This helps with larger articles where the memory limit is a challenge. On the other hand, Stuff method summarizes the entire article at once. For ease, we use Stuff method in this study with the first 5K characters of the article due to the memory limits. %Llama generates slightly different summaries from the same article, every time it is invoked. Therefore we summarize every article 3 times and report average metrics in Section~\ref{sec:evaluate}. 
We use the following prompt template to perform summarization: 
% \begin{verbatim}
    
%     Write a concise summary of the following text 
%     delimited by triple backquotes. Return your 
%     response in bullet points which covers the key
%     points of the text.     
%     ```{text}```
%     BULLET POINT SUMMARY:
    
% \end{verbatim}

\begin{verbatim}
Write a summary of the following text 
delimited by triple backticks. Return 
your response which covers the key points 
of the text.
```{text}```
SUMMARY:    
\end{verbatim}

\noindent \textbf{T5 Small:} For Format Fine-tuning, we experimented with the Google T5 model~\cite{T5} on the news datasets. 
For the T5 model text summarization, we used $pipeline()$ method from Hugging Face API directly as we could use a simple prompt ($"summarize:"$) that T5 model was already trained on. %Since the T5 model generates the same summary for an article every time it is invoked, we summarize every article only once using the T5 model. 

%Langchain enables combining LLMs with other functions and running a sequence of function calls. 

% \begin{table*} [!h]
% \begin{center}
% \begin{tabular}{| c | c | c | c | c | c | c |}
% \hline
%   & ROUGE-1 & ROUGE-2 & ROUGE-L & BLEU & METEOR & BERTScore \\ 
%  \hline
%  \hline
%  Foundation & 0.24 & 0.09 & 0.18 & 0.07 & 0.24 & 0.85 \\ 
%  \hline 
%  Fine-tuned & 0.36 & 0.24 & 0.31  & 0.20 & 0.39 & 0.88 \\
%  \hline
% \end{tabular}
% \end{center}
% \caption{Classical metrics on Newsroom dataset T5 model summaries.}
% \label{tab:rouge_newsroom}
% \vspace{-4mm}
% \end{table*}

\subsection{Examples of Detected Invalid Summaries}
\label{ssec:appendix_invalid_eg}
\textbf{Example 1:} This part of a generated summary where LLM output a code snippet.

\noindent Summary:
\begin{verbatim}
def test_line_break(self):
"""
Test line breaks in a document.
"""
document = Document()
document.add_field(name="key", text="..")
\end{verbatim}

\noindent \textbf{Example 2:} This is a summary example where LLM output the word summary multiple times.

\noindent Summary:
You are going to write a short summary of the information provided
in the sample text. The summary can be no longer than 20 words. You must
be able to summarize the text. In other words, a short summary of
the information presented. In the example, the summary has been
The first summary is presented as text that has been highlighted. The
in between the summary text and the source text of the sample text.
The second summary is presented as a summary table. This table has
original sample text. The summary table contains the highlighted
summary text.
                
\noindent \textbf{Example 3:} This is part of a summary where LLM output adobe reader related text which was unrelated to article. This occurred multiple times in our experiments.

\noindent Summary:
The following files are included within this PDF Package.
Adobe Acrobat 8.0 or later is required to view these files.
If you dont already have Adobe Reader 8.0, please click here to download.
This file contains the following.

\noindent \textbf{Example 4:} This is part of a summary where LLM output garbage and no letters.

\noindent Summary:
-----------------------------------------------------------------------------------------------------------------------------------------------------------

\noindent \textbf{Example 5:} This is part of a summary where LLM output garbage. Heuristic to detect this is multiple equal to ($=$) notations.

\noindent Summary:
\begin{verbatim}
<Project Language=".." AssemblyName=..>
  <Document>
    <Block xmlns="http://.."
      id="129_22000"
      name="MainDocumentPart_8">
      <!-- Add the code snippet ..-->
    </Block>
  </Document>
</Project>
\end{verbatim}

%\subsection{Topic Similarity Details}
%\label{ssec:ts_details} 
%\subsubsection{Examples of Topic Modeling Outputs}
%\label{sssec:topic_modeling_op}
\subsection{Examples of Topic Modeling Outputs}
\label{ssec:appendix_topic_modeling}

\textbf{Example of Topics Generated from NARA documents and their summaries:} We show an example of the topics generated by LDA in Table~\ref{tab:eg_topics}.

\noindent \textbf{Example of an output of topics for a document:} Here in Table~\ref{tab:eg_doc_topics}, we show the output of LDA on a randomly picked document. It shows how the document contains varying amounts of the different topics that are shown in example Table~\ref{tab:eg_topics}.

\subsection{Understanding AlignScore}
\label{ssec:alignscore}
In Figure~\ref{fig:as_ts_pdf}, we look at the distribution of Topic Similarity and Alignscore for comparing articles with their valid vs. invalid summaries. Notice that, for TS, clearly, there is a high probability of high values for valid data, and high probability of very low TS values for invalid data. However, for AlignScore, distribution is similar for valid vs. invalid summaries. One reason for this could be the way chunking and aggregation (via taking the maximum values) is achieved in the AlignScore implementation.

\end{document}